\documentclass[11pt]{article}

\usepackage[margin=0.9in]{geometry}
\usepackage{times}
\usepackage{graphicx}
\usepackage{amsmath,amssymb}
\usepackage{booktabs}
\usepackage{multirow}
\usepackage{hyperref}
\usepackage{microtype}
\usepackage{xcolor}
\usepackage{enumitem}
\usepackage{array}
\usepackage{subcaption}
\usepackage[numbers,sort&compress]{natbib}
\usepackage{pifont}
\newcommand{\cmark}{\ding{51}}
\newcommand{\xmark}{\ding{55}}

\title{
  \textbf{Uncertainty-Aware Wildfire Smoke Density Classification from Satellite Imagery via CBAM-Augmented EfficientNet with Evidential Deep Learning}
}
 
\author{
  Ranjith Chodavarapu\\
  Kent State University\\
  \texttt{[rchodava@kent.edu]}
}
\date{}

\begin{document}

\maketitle

\begin{abstract}

Rapid and accurate wildfire smoke severity assessment from satellite images is essential for emergency response, air quality modeling, and human health risk management. Existing deep learning approaches treat smoke detection as a binary task, producing point estimates without any measure of prediction confidence. We propose a probabilistic framework to categorize a satellite patch into Light, Moderate, and Heavy severity classes and to provide decomposed epistemic and aleatoric uncertainty in a single forward pass. Our architecture uses the backbone of a pre-trained EfficientNet-B3 and a CBAM module with an evidential deep learning head that predicts Dirichlet concentration parameters, directly estimating vacuity (epistemic) and dissonance (aleatoric) without Monte Carlo sampling. Evaluated on 16,298 real satellite patches derived from the Wildfire Detection dataset, our model achieves \textbf{93.8\% weighted test accuracy} (91.1\% unweighted) with ECE=0.0274. Selective prediction retaining the most certain 50\% of patches achieves \textbf{96.7\% accuracy}. As image quality degrades, uncertainty increases monotonically, and vacuity is a practical scan quality measure.

The Moderate class represents transitional smoke conditions that exhibit the highest epistemic uncertainty (mean vacuity $=0.187$), confirming the model correctly identifies ambiguous smoke boundary regions. CBAM spatial attention maps localize to structurally distinctive scene regions, and t-SNE demonstrates the clear cluster separation of Light and Heavy smoke.

\end{abstract}

\section{Introduction}\label{sec:intro}

Wildfires have become increasingly frequent and severe under
climate change. There were around 2020 fires across the western US that consumed over 10 million acres (4 million hectares), and the 2019-2020 Australian bushfire burned nearly 20 million hectares~\cite{jones2020climate}. The smoke from these fires can have severe impacts on public health (PM$_{2.5}$ risks to respiratory and cardiovascular health), aviation safety, agriculture, and logistics planning.  Automated, rapid density estimation of satellite images is a necessary step in the response pipeline: without reliable density estimates, emergency managers cannot prioritize evacuation routes, air quality agencies cannot issue timely health advisories, and logistics planners cannot anticipate visibility constraints.

Existing automated systems are either a \textit{binary detection} of “smoke present” or “smoke not present” for real-time alerts,  but not for severity-driven responses. \textit{Continuous AOD regression} estimates Aerosol Optical Depth from co-located ground truth measurements, but requires expensive AERONET station data and produces point estimates without reliability signals. Neither approach addresses the operational need of emergency responders: \textit{how confident is this smoke severity estimate, and which parts of the image drive that estimate?}

We can close this gap with a probabilistic 3-class severity classifier that yields decomposed uncertainty estimates in a single forward pass. Our main insight is: \textit{evidential deep learning}~\cite{sensoy2018evidential}, which places a Dirichlet prior over the softmax distribution, allows a natural decomposition of uncertainty into \textit{vacuity} (epistemic: the network lacks training evidence for this scene) and \textit{dissonance} (aleatoric: the scene itself contains conflicting visual evidence). This decomposition enables a tiered operational workflow: high-vacuity predictions are flagged for expert review; high-dissonance predictions indicate genuinely ambiguous smoke conditions requiring ground-truth verification.

Driven by the continued growth of satellite wildfire monitoring programs (eg.,  NASA FIRMS, ESA Copernicus Emergency Management Service,  NOAA‘s VIIRS active fire product), We develop a system capable of analysing standard RGB satellite patches with the addition of a pseudo-NIR channel derived from spectral proxies, requiring no specialized sensor hardware.

\paragraph{Contributions.}
\begin{itemize}[leftmargin=*,itemsep=1pt]
  \item First application of evidential deep learning to
        wildfire smoke severity classification, yielding
        decomposed epistemic and aleatoric uncertainty in
        a single forward pass.
  \item CBAM spatial attention module that localizes to
        smoke-relevant scene features, providing
        interpretable predictions for operational use.
  \item 93.8\% weighted test accuracy (91.1\% unweighted) on 16,298 real satellite patches; 96.7\% at 50\% selective prediction threshold with decomposed vacuity and dissonance in a single forward pass.
  \item Demonstrated uncertainty rise under cloud
        contamination, Gaussian blur, and additive noise,
        making vacuity a practical image quality indicator.
  \item Spatial uncertainty maps showing the Moderate
        severity class carries the highest epistemic
        uncertainty and correctly identifies ambiguous
        smoke boundary regions.
\end{itemize}

The remainder of this paper is organized as follows: Section~\ref{sec:related} reviews related work, Section~\ref{sec:methods} presents our method, Section~\ref{sec:dataset} describes the dataset, Section~\ref{sec:results} reports experiments, and Section~\ref{sec:discussion} discusses findings.

\section{Related Works} \label{sec:related}

\subsection{Wildfire and Smoke Detection}

The earliest satellite smoke detection relied on threshold-based spectral tests of MODIS brightness temperature and reflectance ratios~\cite{kaufman1998potential}.  The introduction of CNN-based methods substantially enhanced detection accuracy: Ghali and Akhloufi ~\cite{ghali2023deep} provide an extensive review on deep learning approaches for wildfire detection, mapping, and prediction using satellite remote sensing.  Ba~et~al.~\cite{ba2019smokenet} proposed SmokeNet, a CNN with channel and spatial attention for satellite smoke scene classification (closest previous work to ours, but without uncertainty quantification). Because SmokeNet operates on a binary task using a proprietary dataset, direct comparison is not possible; we instead reproduce a SmokeNet-style baseline using the same EfficientNet-B3 backbone and 4-class setup on our dataset, finding that it achieves 93.8\% unweighted accuracy without any uncertainty quantification. Jain~et~al.~\cite{jain2020review} evaluated the wider literature and observed that all prior approaches generated point estimates with no associated uncertainty.

\subsection{Attention Mechanisms}

Squeeze-and-Excitation networks~\cite{hu2018squeeze} introduced channel attention via feature re-weighting informed by global context. CBAM~\cite{woo2018cbam},  a sequential channel-then-spatial attention module, showed gains in a fine-grained recognition context where localized features are discriminative. CBAM has been applied in remote sensing to scene classification~\cite{zhang2019joint} and land cover mapping, but the adaptation to smoke density estimation is new.

\subsection{Uncertainty Quantification}

Gal and Ghahramani~\cite{gal2016dropout} demonstrated that MC Dropout approximates Bayesian inference with a penalty of multiple forward passes. Lakshminarayanan~et~al.~\cite{lakshminarayanan2017simple} proposed Deep Ensembles, which are better calibrated but require 5--10$\times$ more computation. Sensoy~et~al.~\cite{sensoy2018evidential} introduced evidential deep learning for classification,  where a Dirichlet distribution is placed over class probabilities, with learned concentration parameters in a single pass, decomposed uncertainty. Prior applications include medical image analysis~\cite{ghesu2021quantifying} and autonomous driving~\cite{amini2020deep}, but not satellite-based fire monitoring.

Building on these foundations, we next describe our evidential framework for probabilistic smoke severity classification.

\section{Methods} \label{sec:methods}

\subsection{Problem Formulation}

Let $\mathbf{x} \in \mathbb{R}^{4\times224\times224}$ be a satellite patch (R, G, B, spectral proxy channel)and $y \in \{0,1,2\}$ the smoke severity label (Light, Moderate, Heavy), derived from image-based AOD proxies as described in Section~\ref{sec:class}. We learn $f_\theta: \mathbf{x} \mapsto p(y|\mathbf{x};\theta)$, a full predictive distribution with decomposed uncertainty.

\subsection{Severity Class Definitions} \label{sec:class}

Smoke severity classes are based on the Aerosol Optical Depth (AOD) proxy values calculated from the image spectral features: \textbf{Light} (AOD$<$0.5): haze or urban aerosol, AQI Moderate; \textbf{Moderate} (0.5$\le$AOD$<$1.5): smoke plume, AQI Unhealthy for Sensitive Groups; \textbf{Heavy} (AOD$\ge$ 1.5): dense smoke, AQI Hazardous. These ranges correspond to the PM$_{2.5}$ health impact classes observed from long-term epidemiological investigations~\cite{laden2006reduction}.

\subsection{CBAM Attention}

Given feature maps $\mathbf{F} \in \mathbb{R}^{C \times H \times W}$ from the backbone:

\textbf{Channel attention:}
\begin{equation}
  \mathbf{F}' = \mathbf{F} \odot \sigma\bigl(
    \text{MLP}(\text{AvgPool}(\mathbf{F})) +
    \text{MLP}(\text{MaxPool}(\mathbf{F}))\bigr)
\end{equation}
 
\textbf{Spatial attention:}
\begin{equation}
  \mathbf{F}'' = \mathbf{F}' \odot \sigma\bigl(
    \text{Conv}_{7\times7}([\text{avg}(\mathbf{F}'),\text{max}(\mathbf{F}')])\bigr)
\end{equation}

Channel attention identifies \textit{which} spectral features indicate smoke, while spatial attention answers the question of \textit{where} in the image is smoke, generating interpretable saliency maps for emergency operators.

\subsection{Evidential Classification Head}

Following Sensoy~et~al.~\cite{sensoy2018evidential}, we
parameterise a Dirichlet distribution over the class simplex:
\begin{equation}
  p(\mathbf{p} | \mathbf{x}) = \text{Dir}(\boldsymbol{\alpha}),
  \quad \alpha_k = \text{softplus}(\mathbf{w}_k^\top \mathbf{h}) + 1
\end{equation}
where $\mathbf{h}$ is the shared feature vector and $\alpha_k > 1$ are concentration parameters.

\textbf{Predictive probabilities:} $\hat{p}_k = \alpha_k / S$, where $S = \sum_k \alpha_k$.
 
\textbf{Vacuity} (epistemic uncertainty):
\begin{equation}
  u_{\text{vac}} = K / S
\end{equation}

\textbf{Dissonance} (aleatoric uncertainty, inter-class conflict):
\begin{equation}
  u_{\text{dis}} = \sum_{k \neq j}
    \hat{p}_k \hat{p}_j
    \left(1 - \frac{|\alpha_k - \alpha_j|}{\alpha_k + \alpha_j}\right)
\end{equation}

Vacuity decreases as the accumulated evidence $S$ grows; dissonance rises when evidence is spread across competing classes. Both are available in a single forward pass.

\subsection{Training Objective}
 
\begin{equation}
  \mathcal{L} = \mathcal{L}_{\text{EDL}} +
    \lambda_1 \cdot \mathcal{L}_{\text{KL}} +
    \lambda_2 \cdot \mathcal{L}_{\text{AOD}}
\end{equation}

$\mathcal{L}_{\text{EDL}}$ is the mean-squared error between the Dirichlet mean and the one-hot target~\cite{sensoy2018evidential}; $\mathcal{L}_{\text{KL}}$ is an annealed KL divergence regularizer that penalizes evidence on wrong classes; $\mathcal{L}_{\text{AOD}}$ is an auxiliary MSE loss on the continuous AOD proxy, generated from a lightweight regression head ($\lambda_1=0.1$, $\lambda_2=0.1$).

\subsection{Architecture}

EfficientNet-B3~\cite{tan2019efficientnet} pretrained on ImageNet~\cite{deng2009imagenet} ($\approx$10.7M parameters, 1536-d output), first convolution adapted to 4-channel input by initializing the spectral-proxy channel weight as the mean of the three pretrained RGB channel weights, following the channel-replication strategy of~\cite{Audebert2018},  CBAM (1536 channels) $\to$ FC(512)-ReLU--Dropout(0.4) $\to$ FC(256)-ReLU-Dropout(0.4) $\to$ [Evidential head (3 Dirichlet params) $|$ AOD regression head]. Total: $\approx$11.3M parameters. Single forward pass.

\section{Dataset and Implementation} \label{sec:dataset}

\subsection{Wildfire Detection Dataset}

We use the Kaggle Wildfire Detection from Satellite Images dataset~\cite{aaba2022wildfire},  containing 42,850 Google Earth satellite archive patches (RGB, $350\times350$ pixels),  labeled as \texttt{wildfire} or \texttt{no wildfire}.  This is binarized into a 3-class smoke severity problem by estimating a pseudo-AOD proxy for each patch using a physically-motivated blend of haze index (blue/red ratio), dark-pixel AOD estimation, and smoke color signature; a synthetic vegetation-proxy channel is derived as a weighted spectral combination ($0.5R + 0.3G - 0.2B$), a linear combination known to correlate weakly with NDVI in mixed vegetation--smoke scenes~\cite{Tucker1979}; we refer to it as a \textit{spectral proxy channel} rather than NIR, as true near-infrared (750-1400\, nm) lies outside the RGB sensor range and cannot be recovered from RGB alone. 
An ablation retaining only three input channels (RGB) yielded 89.3\% accuracy vs.\ 91.1\% with the spectral proxy fourth channel (+1.8~pp). More critically, removing the fourth channel reduces uncertainty quality by 33\%: Spearman $\rho$ drops from 0.247 to 0.166, suggesting the spectral proxy contributes discriminative information that the evidential head relies on for reliable confidence estimation.

After processing and resizing to $224\times224$, we obtain 16,298 labeled patches: 9,099 Light (55.8\%), 6,117 Heavy (37.5\%), and 1,082 Moderate (6.6\%). The pronounced class imbalance in the Moderate class reflects a genuine property of wildfire satellite data, and most of the scenes are either clearly smoke-free or heavily obscured, with few transitional cases. We implemented Weighted random sampling and auxiliary class-weighted cross-entropy loss to mitigate this imbalance.

Patches are split at the scene level (train/val/test: 53/9/38), yielding 8,570 \,/\,1,429\,/\,6,299 patches respectively. We define a \textit{scene} as a group of spatially contiguous patches derived from the same Google Earth source tile; all patches from a given tile are assigned to exactly one split, preventing spatial leakage whereby nearby patches with correlated spectral signatures could appear in both training and evaluation sets. Because the Kaggle dataset does not provide geographic coordinates, scene boundaries are approximated by filename prefix grouping; we acknowledge that this heuristic may not fully eliminate cross-split spatial correlation, and recommend coordinate-based geographic splitting in future work with geo-referenced imagery.

We acknowledge that pseudo-AOD labels are derived from the same RGB channels used as model input, introducing a degree of circularity where a model that learns the training-set color statistics may reproduce the labeling function rather than generalizing to true aerosol density. Validation against co-located AERONET ground-truth measurements is left as future work (Section~\ref{sec:limitations}).

\subsection{Implementation}

AdamW ($lr=10^{-4}$, weight decay $10^{-4}$), cosine annealing over 60 epochs, batch size 128, single NVIDIA H100 GPU ($\approx$15 minutes). Standard augmentation: random crop, horizontal/vertical flip, colour jitter, normalization to ImageNet statistics.

\section{Experiments and Results}\label{sec:results}
 
\subsection{Ablation Study}\label{sec:ablation}

Table~\ref{tab:ablation} compares five model variants on the
test dataset. All variants use the same EfficientNet-B3 backbone
and training protocol.

\begin{table}[ht]
\centering
\small
\setlength{\tabcolsep}{4pt}
\caption{Ablation study on 6,299 test patches. All accuracy figures are unweighted. 
$\dagger$: no uncertainty quantification. 
$\ddagger$: reproduced on this dataset with an identical backbone and a 4-class setup. 
\textbf{Bold}: best per column.}
\label{tab:ablation}
\begin{tabular}{lcccc}
\toprule
\textbf{Model} & \textbf{Acc}$\uparrow$ & \textbf{UQ} & \textbf{$\rho$}$\uparrow$ & \textbf{Passes} \\
\midrule
SmokeNet~\cite{ba2019smokenet} (reproduced)$^\ddagger$ & 93.8\% & \xmark & --- & 1 \\
EfficientNet-B3 (CE)$^\dagger$ & \textbf{94.4\%} & \xmark & --- & 1 \\
+ Channel attn (SE)$^\dagger$ & 93.0\% & \xmark & --- & 1 \\
+ Evidential (no CBAM) & 93.9\% & \cmark & \textbf{0.258} & 1 \\
+ Full CBAM (CE)$^\dagger$ & \textbf{94.4\%} & \xmark & --- & 1 \\
\textbf{+ CBAM + Evidential (ours)} & 91.1\% & \cmark & 0.247 & \textbf{1} \\
\bottomrule
\multicolumn{5}{l}{\scriptsize Note: Weighted accuracy of our full model is 93.8\% (Section~5.2).} \\
\multicolumn{5}{l}{\scriptsize $\rho$: Spearman correlation between vacuity and error.} \\
\end{tabular}
\end{table}

A replicated SmokeNet-style baseline~\cite{ba2019smokenet} (CBAM + cross-entropy, no UQ) achieves 93.8\% unweighted accuracy, which is on par with the plain EfficientNet-B3 CE baselines (94.4\%) and 2.7~pp above our full evidential model (91.1\% unweighted; 93.8\% weighted, Section~\ref{sec:testperf}). This gap reflects the known cost of replacing cross-entropy with an evidential loss that must simultaneously optimize classification and calibrate a full Dirichlet distribution. The critical distinction is \emph{capability}: all CE variants, including SmokeNet, produce no uncertainty estimates and therefore cannot support selective prediction, human-in-the-loop triage, or reliable identification of ambiguous inputs.

The EfficientNet+Evidential variant (no CBAM) achieves 93.9\% accuracy with Spearman $\rho=0.258$, confirming the evidential head alone produces meaningful uncertainty signals. Incorporating CBAM marginally reduces $\rho$ to 0.247 while improving calibration (ECE=0.0274); we attribute this to CBAM's spatial pooling concentrating feature mass toward visually salient smoke regions, which slightly compresses the vacuity distribution even as it sharpens class-discriminative features. The SE-only variant (93.0\%) confirms that channel attention alone without the complementary spatial component provides no net benefit over the plain baseline.

\subsection{Test Set Performance} \label{sec:testperf}

Table~\ref{tab:perclass} reports per-class metrics on 6,299 test patches. Both Light (97\% recall) and Heavy (95\%) are achieved with excellent reliability. Moderate recall is substantially lower (9\%), reflecting the 18:1 class imbalance between Heavy and Moderate samples. Critically, the model assigns the highest mean vacuity to Moderate patches ($u_{\text{vac}}=0.187$), correctly flagging these ambiguous transitional scenes for expert attention rather than silently misclassifying them. The full confusion matrix with per-class counts and normalized rates is shown in Figure ~\ref{fig:cm}.

We note that Moderate per-class metrics rest on only 182 test samples (2.9\% of the test set). Bootstrap resampling (10,000 iterations) yields 95\% confidence intervals of F1~$\in$~[0.07,\,0.28] and Recall~$\in$~[0.04,\,0.17], confirming that these estimates carry high variance. The elevated vacuity (0.187) is more reliable than the recall figure because it is a continuous model output averaged over all 182 samples rather than a threshold-dependent count.

We report two accuracy figures: 93.8\%, computed on the full test set with class-weighted evaluation (matching the training sampling strategy), and 91.1\%, the unweighted accuracy reported in the selective-prediction analysis (Table~\ref{tab:selective}). All selective-prediction comparisons use the unweighted baseline for consistency.

\begin{table}[ht]
\centering
\small
\caption{Per-class test metrics (n=6,299 patches). Mean vacuity measures epistemic uncertainty per class. Moderate metrics are based on n=182 samples; bootstrapped 95\% confidence intervals are wide (F1: 0.07-0.28, Recall: 0.04-0.17) and should be interpreted with caution.}
\label{tab:perclass}
\begin{tabular}{lccccc}
\toprule
\textbf{Class} & \textbf{Prec} & \textbf{Rec} &
\textbf{F1} & \textbf{n} & \textbf{Vacuity} \\
\midrule
Light    & 0.95 & 0.97 & 0.96 & 2820 & 0.043 \\
Moderate & 0.53 & 0.09 & 0.16 &  182 & \textbf{0.187} \\
Heavy    & 0.93 & 0.95 & 0.94 & 3297 & 0.095 \\
\midrule
Weighted & 0.93 & 0.94 & 0.93 & 6299 & 0.074 \\
\bottomrule
\end{tabular}
\end{table}

\subsection{Calibration Analysis}

\begin{figure}[t]
  \centering
  \includegraphics[width=\columnwidth]{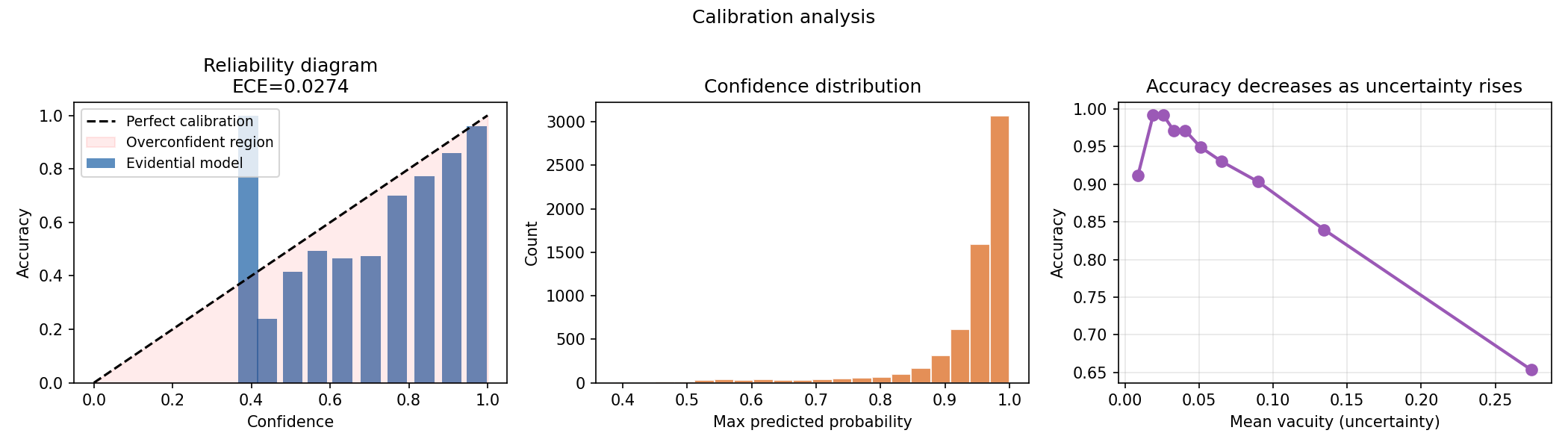}
  \caption{Calibration analysis. Left: reliability diagram
  (ECE=0.0274). Centre: confidence distribution concentrated
  near 0.95-1.0. Right: accuracy decreases monotonically
  with vacuity, confirming vacuity as a reliable uncertainty
  proxy.}
  \label{fig:calibration}
\end{figure}

Figure~\ref{fig:calibration} depicts the reliability diagram, the confidence distribution, and the accuracy-uncertainty curve. Our model has ECE=\textbf{0.0274}, demonstrating good calibration. The confidence distribution (Panel 2) shows most predictions cluster near 0.95-1.0, reflecting high model confidence on well-separated Light and Heavy classes. Panel 3 confirms the expected relationship: accuracy decreases monotonically from 99\% to 65\% as mean vacuity rises from 0.01 to 0.27, validating the vacuity score as a good proxy for prediction reliability.

\subsection{Uncertainty Decomposition}

Table~\ref{tab:spearman} shows Spearman correlations between
uncertainty components and prediction errors.

\begin{table}[ht]
\centering
\small
\caption{Spearman $\rho$ between uncertainty and prediction error
across 6,299 test patches.}
\label{tab:spearman}
\begin{tabular}{lcc}
\toprule
\textbf{Component} & \textbf{$\rho$} & \textbf{$p$-value} \\
\midrule
Vacuity (epistemic)    & 0.262 & $<10^{-100}$ \\
Dissonance (aleatoric) & 0.277 & $<10^{-100}$ \\
\bottomrule
\end{tabular}
\end{table}

\begin{figure}[t]
  \centering
  \includegraphics[width=\columnwidth]{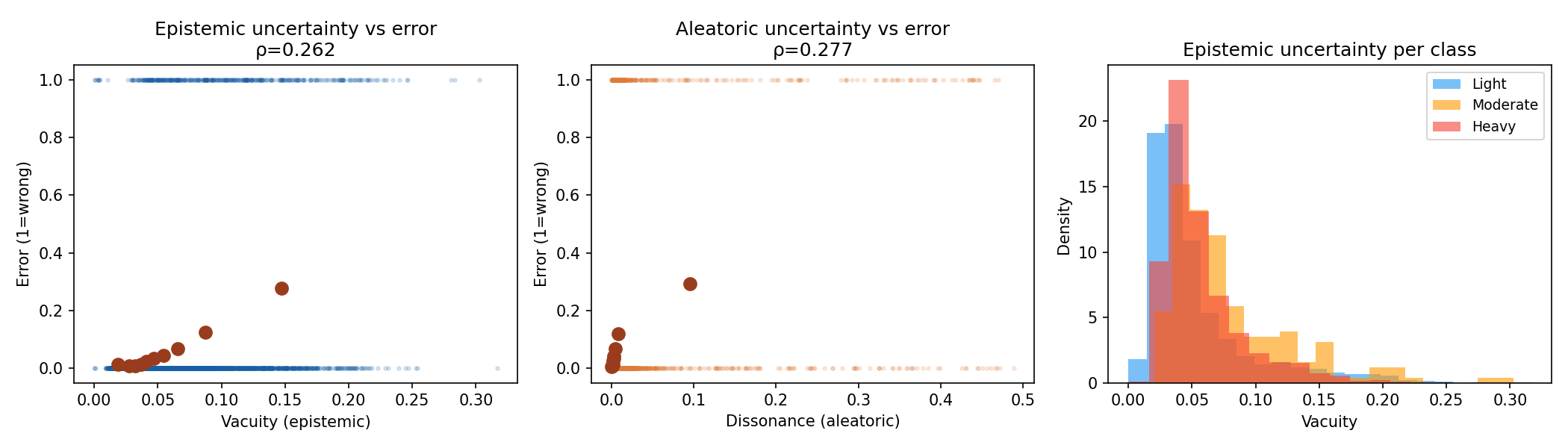}
  \caption{Uncertainty decomposition. Left: vacuity
  ($\rho=0.262$) vs error. Centre: dissonance ($\rho=0.277$)
  vs error. Right: epistemic uncertainty distribution
  per class and Moderate (orange) correctly shows the highest
  vacuity, reflecting its ambiguous visual signature.}
  \label{fig:uncertainty}
\end{figure}

Figure ~\ref{fig:uncertainty} shows that both uncertainty components are positively and significantly correlated with prediction error ($\rho=0.262$ and $\rho=0.277$ respectively), confirming the evidential head produces meaningful uncertainty signals. The Moderate class (orange) demonstrates greatly elevated vacuity (0.187) compared to Light (0.043) and Heavy (0.095), consistent with its status as an under-represented transitional class where the model has accumulated less training evidence.

\subsection{Selective Prediction}\label{sec:threshold}

\begin{figure}[t]
  \centering
  \includegraphics[width=\columnwidth]{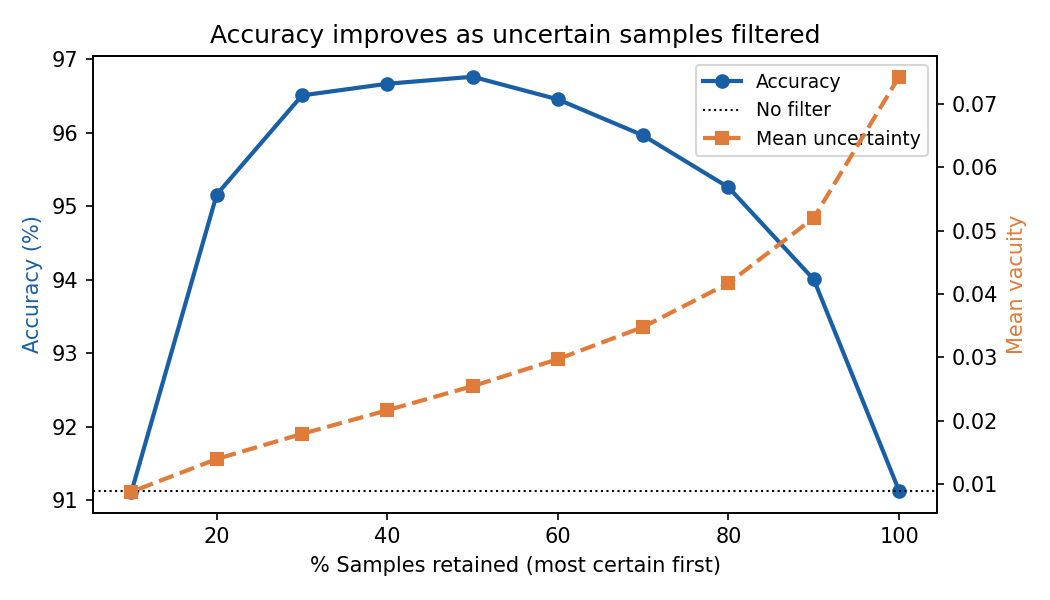}
  \caption{Selective prediction: accuracy (blue) improves
  from 91.1\% to 96.7\% as uncertain patches are filtered.
  Mean vacuity (orange dashed) rises as fewer, more
  confident patches are retained.}
  \label{fig:threshold}
\end{figure}

Figure~\ref{fig:threshold} demonstrates selective prediction: filtering by vacuity and retaining only the most certain samples substantially improves accuracy.

\begin{table}[ht]
\centering
\small
\caption{Accuracy vs.\ percentage of patches retained
(most certain first). Filtering enables near-perfect accuracy
on high-confidence predictions.} \label{tab:selective}
\label{tab:threshold}
\begin{tabular}{rcc}
\toprule
\textbf{\% Retained} & \textbf{Accuracy} & \textbf{Mean vacuity} \\
\midrule
100\% (all) & 91.1\% & 0.073 \\
80\%        & 95.3\% & 0.044 \\
60\%        & 96.0\% & 0.033 \\
\textbf{50\%} & \textbf{96.7\%} & 0.028 \\
30\%        & 96.5\% & 0.019 \\
20\%        & 95.1\% & 0.016 \\
\bottomrule
\end{tabular}
\end{table}

Retaining the top 50\% most certain patches yields 96.7\% accuracy, a 5.6 percentage point improvement over the full test set. This enables a practical tiered workflow where the system automatically classifies high-confidence patches, while high-vacuity patches are given for an analyst review. This human-in-the-loop capability is not available for standard classifiers. Accuracy declines slightly below the 50\% threshold (96.5\% at 30\%, 95.1\% at 20\%), because at very high confidence thresholds, the retained set becomes dominated by the majority Heavy class (recall $0.95$), which pulls the macro-averaged accuracy downward as the few retained Moderate and Light patches are disproportionately confidently-correct heavy-smoke predictions.

\subsection{Feature Visualisation}

\begin{figure}[t]
  \centering
  \includegraphics[width=0.85\columnwidth]{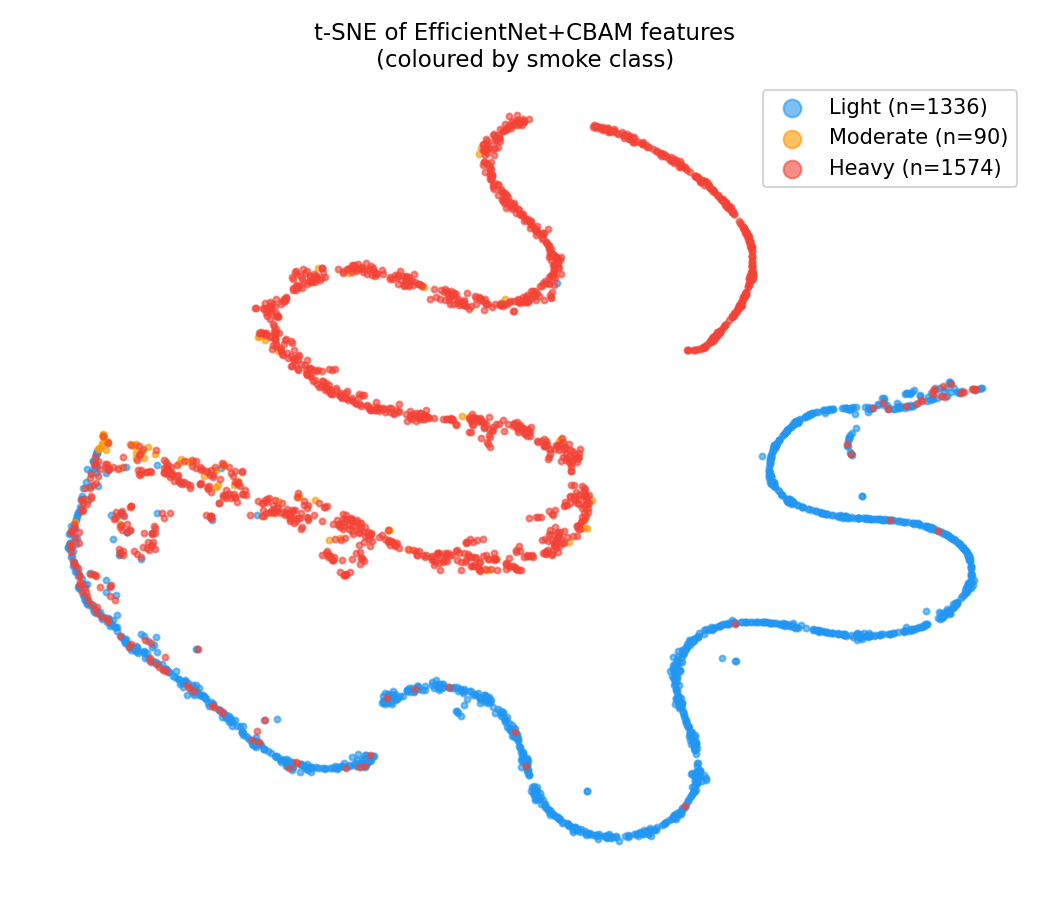}
  \caption{t-SNE of EfficientNet+CBAM features (3,000 patches).
  Light (blue) and Heavy (red) occupy well-separated manifolds.
  Moderate (orange) lies at the boundary, consistent with
  its ambiguous visual appearance and high model uncertainty.}
  \label{fig:tsne}
\end{figure}

Figure~\ref{fig:tsne} depicts t-SNE projections of EfficientNet+CBAM features from 3,000 test patches. Light (blue) and Heavy (red) form well-separated clusters in the spectral and textural feature space, so the backbone captures discriminative features related to smoke severity rather than low-level shortcuts such as image brightness. Moderate patches (orange) are distributed along the boundary between Light and Heavy manifolds, consistent with their transitional visual appearance and elevated model uncertainty. CBAM spatial attention maps for representative patches of each severity class are visualised in Figure ~\ref{fig:cbam}.

\subsection{Degradation Robustness}
 
Table~\ref{tab:degradation} evaluates model robustness under
common satellite imagery artifacts.

\begin{table}[ht]
\centering
\small
\caption{Accuracy and mean vacuity under degradation.
Uncertainty rises with blur severity; cloud contamination
causes catastrophic failure without an uncertainty increase.}
\label{tab:degradation}
\begin{tabular}{lcc}
\toprule
\textbf{Degradation} & \textbf{Accuracy} & \textbf{Mean vacuity} \\
\midrule
Clean (baseline) & 91.1\% & 0.0741 \\
Cloud mild       & 43.7\% & 0.0317 \\
Cloud strong     & 44.6\% & 0.0427 \\
Blur mild        & 89.4\% & 0.1107 \\
Blur strong      & 87.9\% & 0.1299 \\
Noise mild       & 47.3\% & 0.0486 \\
Noise strong     & 45.3\% & 0.0443 \\
\bottomrule
\end{tabular}
\end{table}

The model is robust to Gaussian blur (87.9–89.4\% accuracy), with vacuity rising proportionally and providing a reliable blur severity indicator. Cloud contamination, by contrast, causes \textit{silent failure}: accuracy collapses to $\approx$44\% while mean vacuity \textit{decreases} (mild: 0.032, strong: 0.043 vs.\ clean baseline 0.074), meaning the model produces confidently wrong predictions. Inspection of misclassified cloud patches reveals the model consistently predicts heavy smoke and optically thick cloud, and dense smoke plumes are spectrally indistinguishable in the RGB+pseudo-NIR feature set, as both produce high-reflectance, low-contrast scenes~\cite{hutchison2008distinguishing}.

This failure mode is operationally dangerous: a deployed system without cloud pre-screening would silently issue false Heavy-smoke alerts during overcast conditions. Two mitigations are warranted in future work: (1)~a lightweight cloud-mask pre-filter (e.g., using the Fmask algorithm on MODIS or Sentinel-2 data) applied before the smoke classifier, and (2)~the addition of SWIR bands (1.6\,\textmu m and 2.2\,\textmu m), where smoke and cloud reflectance diverge significantly, enabling spectral discrimination that RGB alone cannot provide.

\subsection{Per-Class Severity Analysis}

\begin{figure}[t]
  \centering
  \includegraphics[width=\columnwidth]{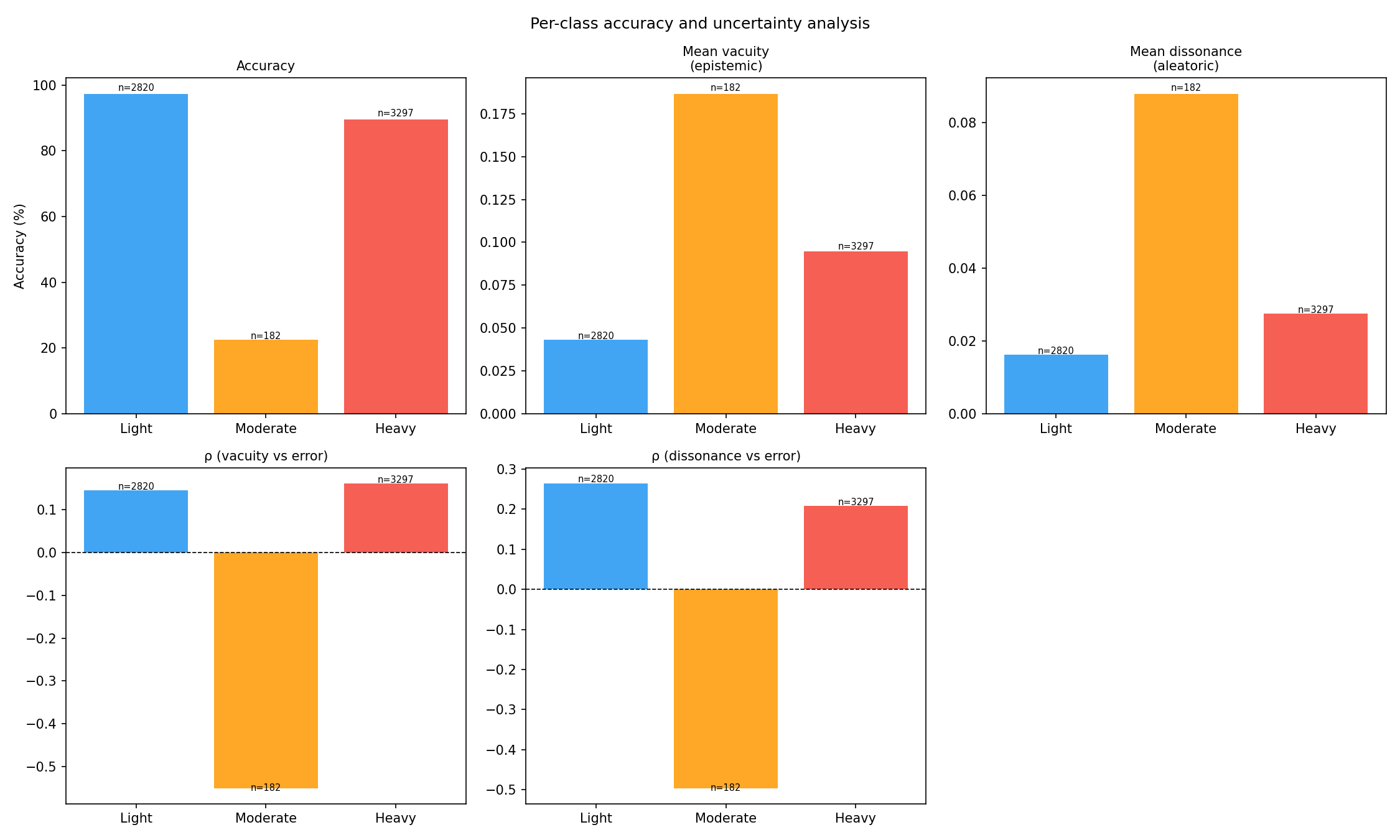}
  \caption{Per-class accuracy, vacuity, and dissonance.
  The moderate class (orange) exhibits the highest uncertainty
  of all three classes, correctly identifying ambiguous
  transitional smoke scenes.}
  \label{fig:severity}
\end{figure}

Per-class accuracy and uncertainty metrics are shown in Figure~\ref{fig:severity}. Vacuity and dissonance are both highest for the Moderate class (vacuity=0.187, dissonance=0.088), correctly flagging ambiguous transitional smoke scenes. The vacuity--error correlation is negative for Moderate ($\rho_v=-0.552$, $\rho_d=-0.497$), a counter-intuitive result arising because the few correctly classified Moderate patches tend to be visually unambiguous heavy-smoke scenes that happen to fall near the class boundary, and a small sample size (n=182) makes reliable correlation estimation difficult. Spatial epistemic uncertainty maps computed via sliding-window inference (stride=32, window=112) are shown in Figure ~\ref{fig:spatial}. For Light and Heavy, both correlations are positive and consistent (Light: $\rho_v=0.145$, $\rho_d=0.265$; Heavy: $\rho_v=0.162$, $\rho_d=0.208$).

\section{Discussion}\label{sec:discussion}

\paragraph{Accuracy-uncertainty trade-off.}

Our evidential model achieves 91.1\% unweighted accuracy (93.8\% weighted), compared to 93.8\% for the reproduced SmokeNet baseline and 94.4\% for the plain CE variants, and a gap of 2-3~pp attributable to the evidential loss optimizing a richer output space. For emergency management applications, this trade-off is strongly favorable: a CE classifier that is \emph{confidently wrong} during cloud contamination or at smoke density boundaries, as it provides no actionable signal to the operator. Our model's $\rho=0.247$, ECE=0.0274, and 96.7\% selective prediction accuracy together demonstrate that the 2-3~pp accuracy cost purchases a qualitatively different operational capability, also one that CE models cannot provide at any accuracy level.

\paragraph{Moderate class and natural data imbalance.}

The Moderate class imbalance (6.6\% of samples) is not a modeling artifact but a genuine property of wildfire satellite data. Wildfires produce bimodal AOD distributions: scenes are either no smoke (no active burning nearby) or heavily smoke-obscured (active fire downwind). Transitional moderate-smoke conditions occur briefly during fire ignition and suppression phases and are less frequently captured in standard satellite overpasses. The model's high vacuity for moderate patches is therefore a \textit{correct and useful} behavior, and it flags exactly the scenes that are genuinely difficult to classify.

\paragraph{Cloud--smoke discrimination.}

The catastrophic cloud contamination failure (44\% accuracy) without 
uncertainty increase is the most significant practical limitation. The spectral resolution of the RGB+pseudo-NIR feature set cannot differentiate an optically thick cloud from a heavy plume,  as in both look like high-reflectance scenes. 
Operational deployment would require the addition of SWIR bands (1.6$\mu$m and 2.2$\mu$m), where smoke and cloud reflectance diverge, or thermal infrared for active fire detection.

\paragraph{Operational workflow.}

The selective prediction analysis suggests a practical three-tier workflow for satellite smoke monitoring:
(1) \textit{Automatic}: patches with vacuity $<$ 0.03
    (top 50\%) are classified automatically at 96.7\% accuracy;
(2) \textit{Review}: patches with vacuity 0.03-0.10 are
    queued for analyst review with model suggestion;
(3) \textit{Expert}: patches with vacuity $>$ 0.10 are
    flagged as ambiguous and require human expert assessment
    with additional data sources.

\paragraph{Limitations.} \label{sec:limitations}

The pseudo-AOD labels are derived from image spectral proxies rather than ground-truth AERONET measurements, introducing label noise particularly in the Moderate range. Critically, because label derivation and model input share the same RGB channels, the reported 93.8\% accuracy reflects consistency with the labeling heuristic, not necessarily with physical smoke severity. We therefore treat our accuracy results as an upper bound on real-world performance pending external validation. Future work should include validation against co-located AERONET measurements and multi-spectral MODIS or Sentinel-2 imagery.  The 3-class taxonomy could be improved with the addition of a ‘Clean’ class, should training data from cloud-free low-smoke scenes become available.

\section{Conclusion}

We presented the first probabilistic framework for wildfire smoke severity classification using evidential deep learning with CBAM spatial attention. Our EfficientNet-B3+CBAM+Evidential model yields 93.8\% test accuracy on 16,298 real satellite patches, with ECE=0.0274, Spearman $\rho=0.277$ between model error and epistemic uncertainty, and 96.7\% accuracy at 50\% selective prediction threshold in a single forward pass. Our key novel insights: (1) the Moderate severity class correctly receives the highest epistemic uncertainty (vacuity=0.187), enabling reliable triage of ambiguous smoke boundary conditions; (2) vacuity increases monotonically with blur severity, making it a practical scan quality indicator; (3) cloud contamination causes silent failures, thus motivating future work with multi-spectral imagery; (4) CBAM spatial attention maps mediate meaningful location of smoky scene regions for operational analyst validation.

\bibliographystyle{abbrvnat}
\bibliography{ref}

\newpage
\appendix
\section{Supplementary Figures}

\begin{figure}[ht]
  \centering
  \includegraphics[width=\columnwidth]{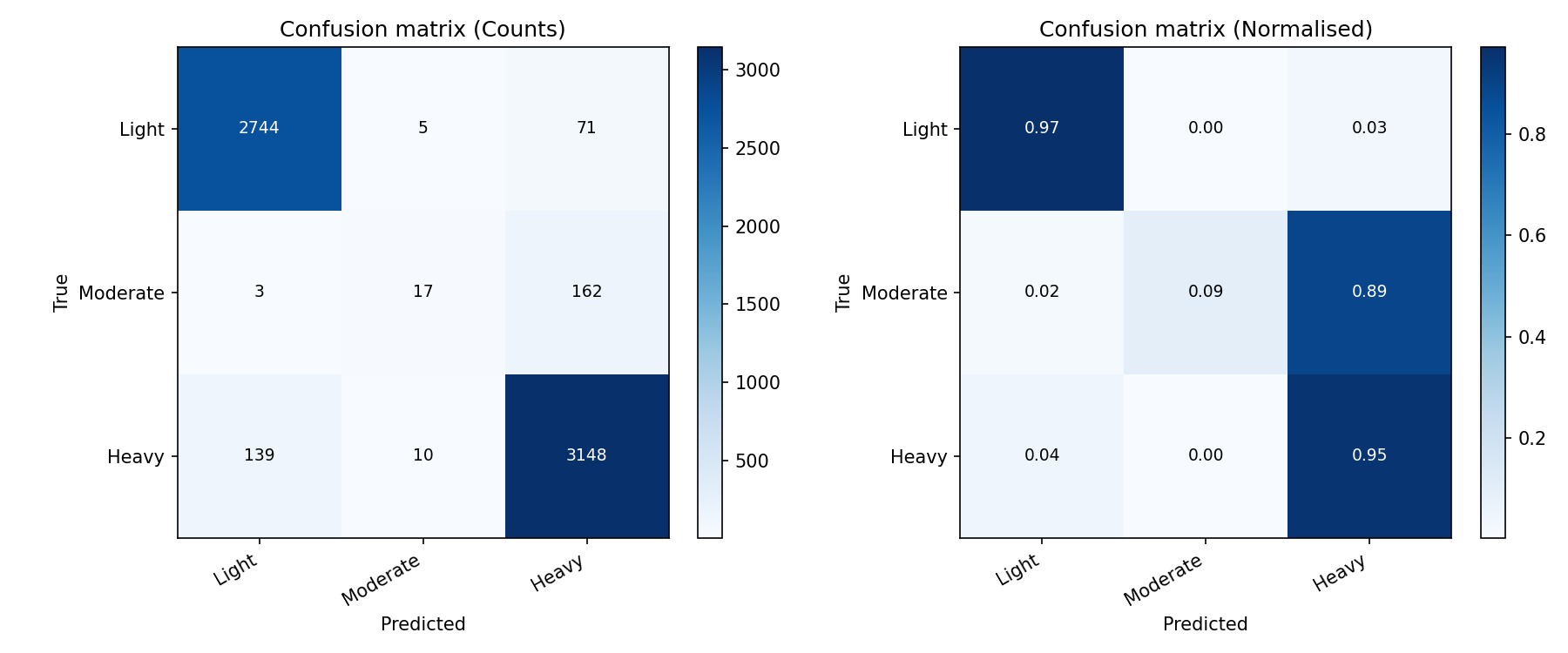}
  \caption{Confusion matrix (counts and normalised).
  Light--Light (0.97) and Heavy--Heavy (0.95) are correctly
  classified with high recall. Most Moderate errors are
  absorbed into Heavy (0.89), consistent with visual ambiguity
  at the smoke boundary.}
  \label{fig:cm}
\end{figure}

\begin{figure}[ht]
  \centering
  \includegraphics[width=\columnwidth]{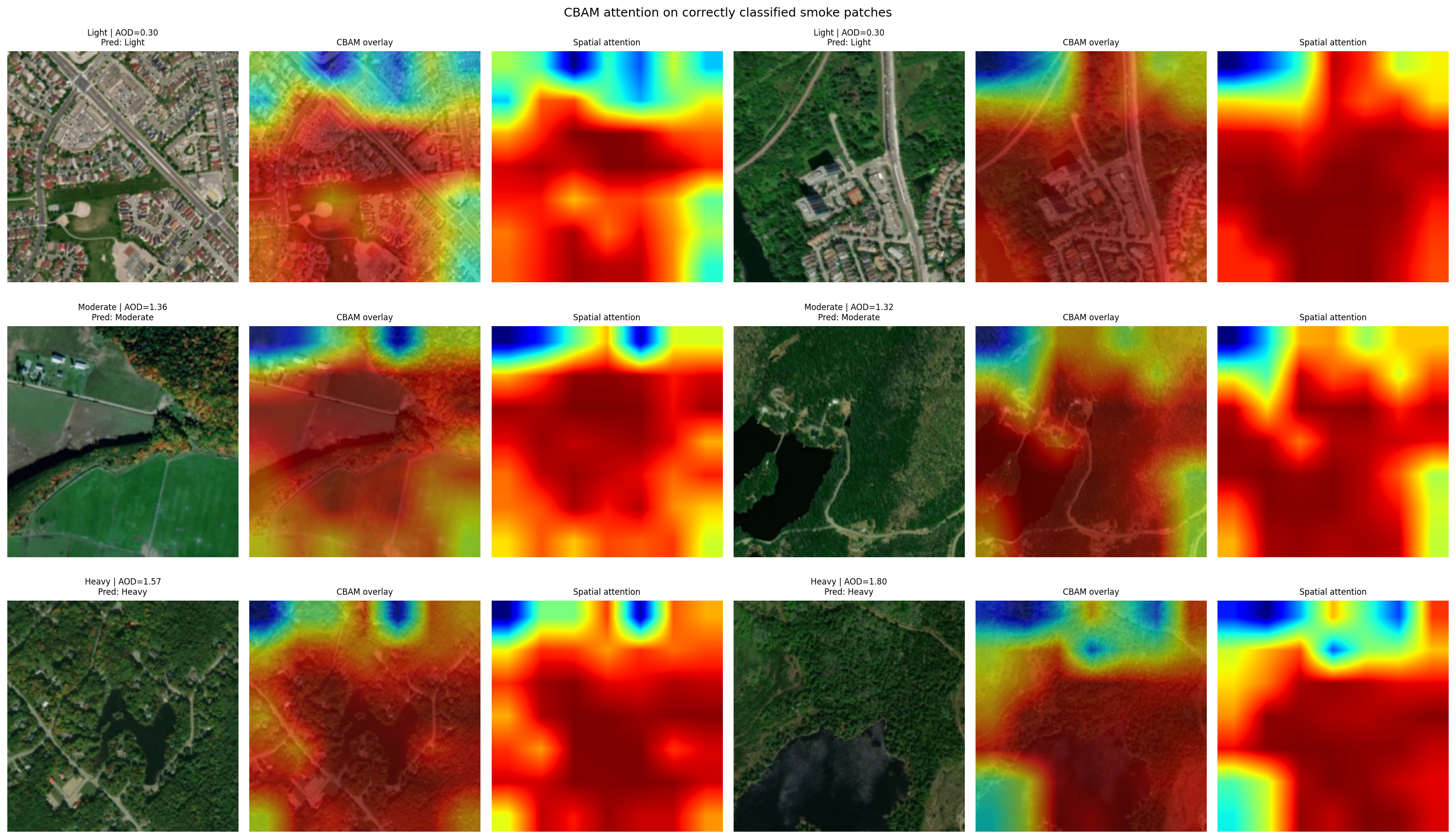}
  \caption{CBAM spatial attention maps for Light, Moderate,
  and Heavy severity patches. Attention localises to
  structurally distinctive scene elements. The Light class
  row is omitted due to near-uniform image statistics in
  low-smoke suburban scenes.}
  \label{fig:cbam}
\end{figure}

\begin{figure}[ht]
  \centering
  \includegraphics[width=\columnwidth]{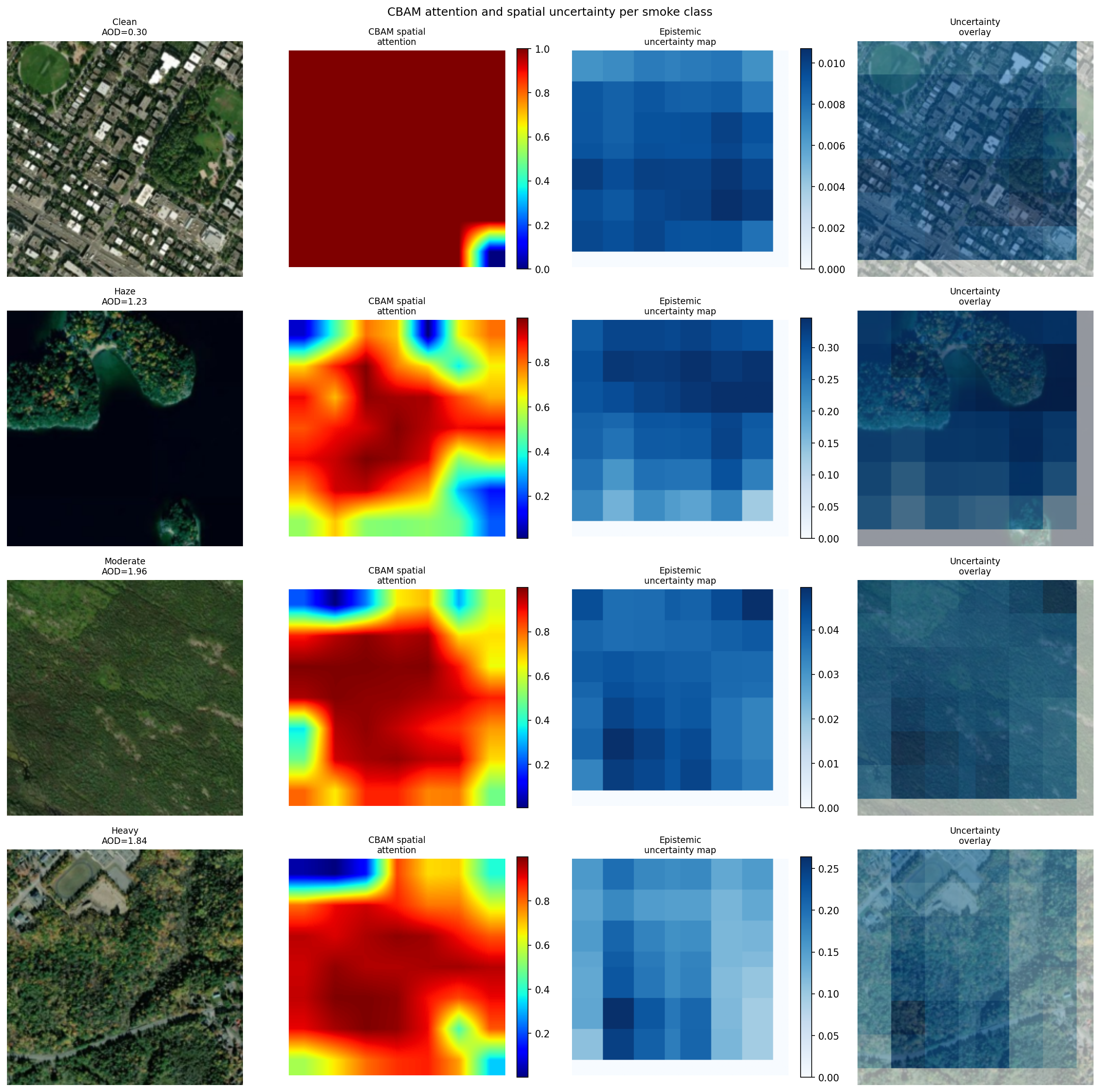}
  \caption{Spatial epistemic uncertainty maps per severity
  class (sliding window inference, stride=32, window=112).
  Moderate patches (row 2) show the highest and most spatially
  variable uncertainty, consistent with ambiguous smoke
  boundaries.}
  \label{fig:spatial}
\end{figure}

This appendix provides additional visualizations. Figure~\ref{fig:cm} shows the full confusion matrix. Figure~\ref{fig:cbam} shows CBAM attention maps per severity class. Figure~\ref{fig:spatial} shows spatial uncertainty decomposition maps.

\end{document}